\documentclass[runningheads]{llncs}
\usepackage{graphicx}

\usepackage{tikz}
\usepackage{comment}
\usepackage{amsmath,amssymb} 
\usepackage{color}
\usepackage{booktabs}
\usepackage{caption,subcaption}
\usepackage{multirow}
\usepackage[bottom]{footmisc}

\usepackage[accsupp]{axessibility}  

\begin{document}
\pagestyle{headings}
\mainmatter
\def\ECCVSubNumber{2021}  

\title{Adaptive Patch Exiting for Scalable Single Image Super-Resolution} 

\titlerunning{Adaptive Patch Exiting (APE)}
%
\author{Shizun Wang\inst{1}\thanks{Equal Contribution} \and
Jiaming Liu\inst{2,4*} \and
Kaixin Chen\inst{1} \and
Xiaoqi Li\inst{2,4} \and \\
Ming Lu\inst{3}\thanks{Corresponding Author} \and
Yandong Guo\inst{4}
}
\authorrunning{S. Wang et al.}
%
\institute{Beijing University of Posts and Telecommunications \and
Peking University \and
Intel Labs China \and 
OPPO Research Institute\\
\email{wangshizun@bupt.edu.cn, lu199192@gmail.com}}
\maketitle

\begin{abstract}
Since the future of computing is heterogeneous, scalability is a crucial problem for single image super-resolution. Recent works try to train one network, which can be deployed on platforms with different capacities. However, they rely on the pixel-wise sparse convolution, which is not hardware-friendly and achieves limited practical speedup. As image can be divided into patches, which have various restoration difficulties, we present a scalable method based on Adaptive Patch Exiting (APE) to achieve more practical speedup. Specifically, we propose to train a regressor to predict the incremental capacity of each layer for the patch. Once the incremental capacity is below the threshold, the patch can exit at the specific layer. Our method can easily adjust the trade-off between performance and efficiency by changing the threshold of incremental capacity. Furthermore, we propose a novel strategy to enable the network training of our method. We conduct extensive experiments across various backbones, datasets and scaling factors to demonstrate the advantages of our method. Code is available at \url{https://github.com/littlepure2333/APE}.
\end{abstract}

\section{Introduction}
\label{sec:intro}

Super-Resolution (SR) is an important technique and has been widely used in video compression \cite{khani2021efficient}, rendering acceleration \cite{xiao2020neural}, network streaming \cite{yeo2020nemo}, medical imaging \cite{sui2020learning}, computational photography \cite{wronski2019handheld} and so on. As the development of Deep Neural Networks (DNNs), plenty of DNN-based methods are proposed for Single Image Super-Resolution (SISR) \cite{shi2016real,dong2014learning,lim2017enhanced,zhang2018image,kim2016accurate,li2021lapar,zhang2021edge}. Existing methods mostly cascade convolutional layers many times to construct deep networks and adopt the pixel-shuffle layer \cite{shi2016real} to obtain high-resolution output. The cascaded layers increase the network's capacity of modeling contextual information over larger image regions. Although significant improvements have been made in performance or efficiency over the past few years, the trade-off between performance and efficiency is still under-explored to the best of our knowledge.

Since there are various hardware platforms like CPUs, GPUs, FPGAs and so on, training one scalable network that can be deployed on platforms with different capacities is strongly demanded for future heterogeneous computing. Recently, a pixel-wise adaptive inference method for scalable SISR has been proposed \cite{liu2020deep}. It learns a predictor to generate the pixel-wise depth map that indicates the target number of layers for each pixel. Sparse convolution guided by the pixel-wise depth map is implemented to achieve speedup. The scalability of \cite{liu2020deep} is realized by changing the mean average of layers for all pixels. However, although \cite{liu2020deep} can obtain theoretical FLOPs reduction, the practical speedup is limited since the pixel-wise sparse convolution is not hardware-friendly. Inspired by the fact that image can be divided into patches, which have various restoration difficulties, \cite{kong2021classsr} proposes a general framework that applies appropriate networks to different patches. A module is learned to classify the patches into various restoration difficulties. They train several models with different capacities to super-resolve patches with different difficulties. Although \cite{kong2021classsr} can save up to 50\% FLOPS on benchmarking datasets, we observe it has two limitations. Firstly, it applies one fixed network to a certain restoration difficulty, which cannot adjust the trade-off between performance and efficiency as \cite{liu2020deep}. Secondly, it needs to store one network for each restoration difficulty, heavily increasing the model size.

To solve the above limitations, we present a scalable method based on Adaptive Patch Exiting (APE) for SISR. Our method can train one network to adaptively super-resolve patches with different difficulties. To be more specific, we train a regressor to predict the incremental capacity of each layer for the input patch. The incremental capacity can evaluate the necessity of each layer. Once the incremental capacity is below a threshold, the patch can exit at the specific layer. Our method can easily adjust the trade-off between performance and efficiency by changing the threshold of incremental capacity. On platforms with high computational resources, our method can lower the threshold to utilize more layers for super-resolution. On platforms with low computational resources, our method can raise the threshold to make the patches exit earlier. Therefore, our method is scalable over platforms with different computational resources. Compared with \cite{kong2021classsr}, which classifies the patches into certain restoration difficulties, our method enables the scalability by adjusting the threshold. In addition, our method only needs to store one network for all restoration difficulties, significantly reducing the model size.

In order to enable the network training, we further propose a strategy that can jointly train the regressor and SR network. Our strategy first train the multi-exit SR network based on the original network, then we calculate the target incremental capacity of each layer based on the multi-exit SR network. Finally, we jointly train the SR network and regressor to converge.

Our contributions can be concluded as follows:
\begin{itemize}
\item We present a novel scalable method for SISR based on adaptive patch exiting, which can be deployed on platforms with different capacities.
\item We propose to learn the incremental capacity of each layer instead of patch difficulty, enabling the patch to exit at the optimal layer.
\item We introduce an effective joint training strategy to enable the training of incremental capacity regressor and SR network.
\item We conduct detailed experiments across various SR backbones and scaling factors to demonstrate the advantages of our method over existing approaches.
\end{itemize}

\section{Related Work}

\textbf{Single Image Super-Resolution} Since the seminal work SRCNN \cite{dong2014learning}, which first applies DNN to SISR, many methods have been proposed. For example, VDSR \cite{kim2016accurate} adopts a very deep neural network to learn the image residual. EDSR \cite{lim2017enhanced} analyzes the DNN layers and proposes to remove some redundant layers from SRResNet \cite{ledig2017photo}. RDN \cite{zhang2018residual} uses dense connections that fully utilize the information of preceding layers. RCAN \cite{zhang2018image} explores the attention mechanism and proposes attentive DNNs to boost the performance. In order to reduce the computational cost, FSRCNN \cite{dong2016accelerating} and ESPCN \cite{shi2016real} propose to use LR image as input and upscale the feature map at the end of networks. LAPAR \cite{li2020lapar} presents a method based on linearly-assembled pixel-adaptive regression network, which learns the pixel-wise filter kernel. In addition to methods focusing on network design, many works study the real-world SR problem. RealSR \cite{cai2019toward} builds a real-world dataset with paired LR-HR images captured by adjusting the focal length. They also present a Laplacian pyramid-based kernel prediction network to recover the HR image. Zero-Shot SR \cite{shocher2018zero} exploits the power of DNN without relying on prior training. They train a small image-specific DNN at test time on examples extracted from the input image itself. Recently, the community also shows the trend of applying techniques like network pruning, quantization, distillation, AutoML to SR. BSRN \cite{xin2020binarized} designs a bit-accumulation mechanism to approximate the full-precision convolution with a value accumulation scheme. Although plenty of DNN-based methods are proposed to improve the performance or efficiency, the scalability problem is still under-explored to the best of our knowledge.

\textbf{Adaptive Inference} Since the future of computing is heterogeneous, training one scalable network that can be deployed on platforms with different capacities is a very important problem. \cite{yu2018slimmable} proposes a simple method that trains a single network executable at different widths, enabling instant and adaptive performance-efficiency trade-off at runtime. \cite{yu2019universally} further extends the slimmable networks \cite{yu2018slimmable} from a predefined widths set to arbitrary width, and generalizes to networks both with and without batch normalization layers. \cite{yu2019autoslim} presents a method that trains a single slimmable network to approximate the network performance of different channel configurations, and then searches the optimized channel configurations under different resource constraints. Instead of switching network width, \cite{jin2020adabits} investigates the option that achieves instant and flexible deployment by adaptive bit-widths of weights and activations in the model. \cite{yang2020mutualnet} trains a set of sub-networks with different widths using different input resolutions to mutually learn multi-scale representations for each sub-network. The performance-efficiency trade-off can be achieved by changing both the network width and input resolution. \cite{cai2019once} proposes to train a once-for-all network that supports diverse platforms by decoupling training and search. They can quickly get a specialized sub-network by selecting from the once-for-all network without additional training. Although plenty of methods are proposed for adaptive inference, they mainly focus on high-level vision tasks. The scalability problem of low-level vision tasks is still under-explored as far as we know. Inspired by the fact that different image regions have different restoration difficulties, \cite{liu2020deep} introduces a lightweight adapter module, which takes image features and resource constraints as input and predicts a pixel-wise depth map. Therefore, only a fraction of the layers in the backbone is performed at a given position according to the predicted depth. While \cite{liu2020deep} can achieve theoretical FLOPS reduction, the practical speed gain is limited since unstructured sparse convolution is not hardware friendly. \cite{kong2021classsr} also utilizes the properties of different image regions by dividing the images into local patches. They train a module to classify the patches into different difficulties, and apply appropriate model to each difficulty. Although \cite{kong2021classsr} can obtain practical speed gains, it is not scalable under different resource constraints.

\section{Method}

\begin{figure*}[t]
	\centering
	\includegraphics[width=0.98\textwidth]{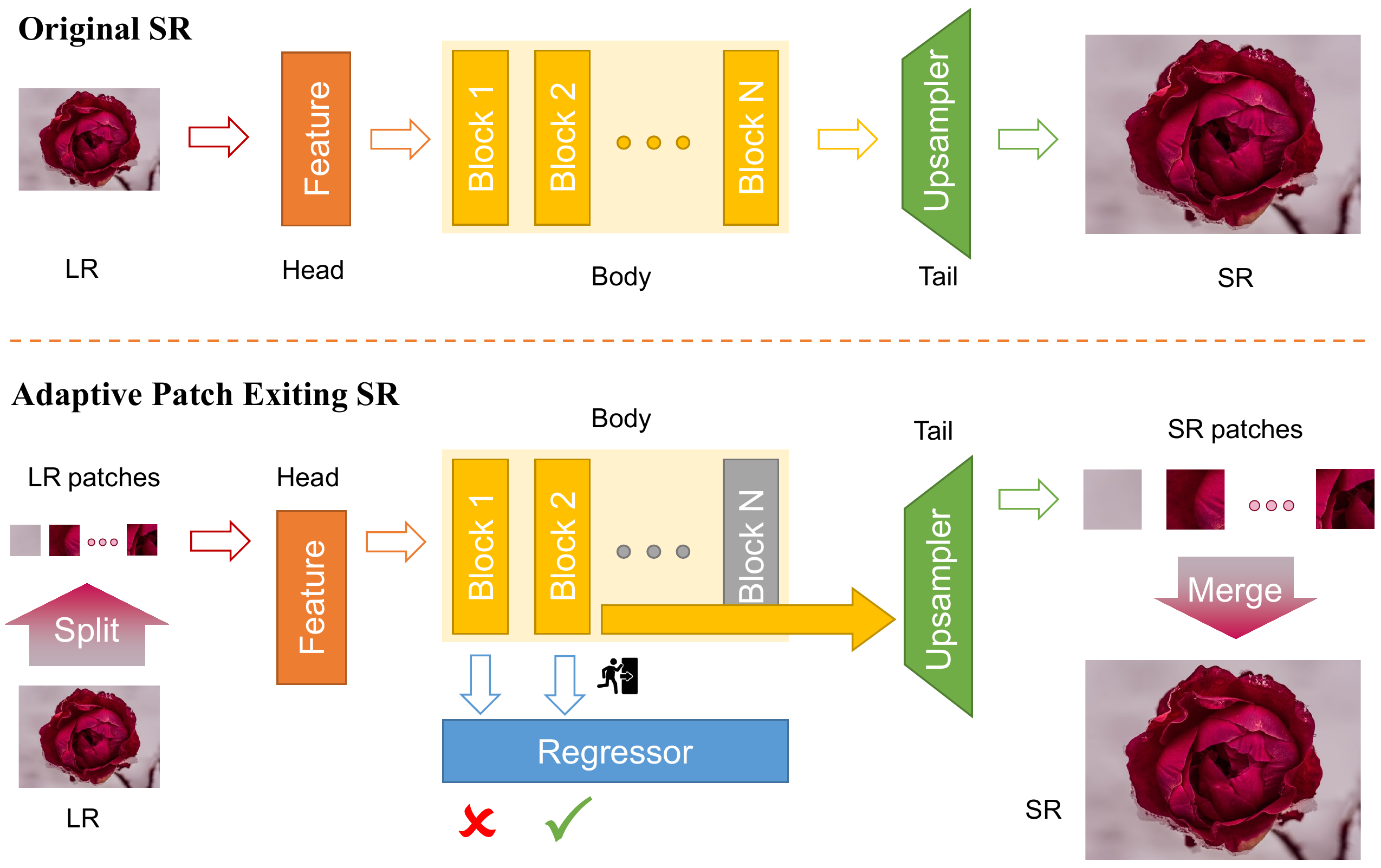}
	\caption{The pipeline of Adaptive Patch Exiting (APE). Original SR networks take a LR image as input, and forward it through head, body and tail to generate the SR image. Instead, APE first splits the LR image into patches, which are forwarded in parallel. The patches will exit early if the incremental capacity estimated by the regressor is below a given threshold. Finally, the SR patches are merged to the output image.}
	\label{fig:pipeline}
\end{figure*}

To apply Adaptive Patch Exiting (APE) to existing SR
networks, we modify the original SR networks to multi-exit
networks and present the training strategy in Sec. \ref{sec:multi_exit_sr}. We then analyze the performance of each patch at a certain layer, and introduce the incremental capacity to evaluate the necessity of each layer for a patch in Sec. \ref{sec:estimate}. Finally, we jointly train the SR network and lightweight regressor in Sec. \ref{sec:ape}. The regressor is used to estimate the incremental capacity at a certain layer. The trained network can achieve the trade-off between performance and efficiency by adjusting the threshold of incremental capacity. The overall pipeline is illustrated in Fig. \ref{fig:pipeline}.

\subsection{Training Multi-Exit SR Networks}
\label{sec:multi_exit_sr}

Super-resolution aims to recover a High-Resolution (HR) image $\hat{y}$ from a given Low-Resolution (LR) image $x$. Since the pioneering work \cite{dong2014learning}, most of the SR networks consist of three parts: head, body and tail. The head part $H$ extracts the features $f_0$ from the LR image:

\begin{equation}
f_0 = H(x;\Theta_h)
\end{equation}
\label{equ:body}

and the body part $B$ enhances $f_0$ by cascading $N$ convolutional layers to generate feature $f_N$:
\begin{equation}
f_N = B(f_0;\Theta_b)
\end{equation}

Finally, the tail part $T$ takes the enhanced feature $f_N$ to obtain the SR output $\hat{y}$:
\begin{equation}
\hat{y} = T(f_N;\Theta_t)
\end{equation}

where $\Theta_h$, $\Theta_b$ and $\Theta_t$ denote the parameters of head, body and tail individually.
Typical SISR architectures such as EDSR \cite{lim2017enhanced}, RCAN \cite{zhang2018image}, VDSR \cite{kim2016accurate} and ECBSR \cite{zhang2021edge} all follow this pipeline. We denote the original SR network as $F$:
\begin{equation}
\hat{y} = F(x;\Theta_h,\Theta_b,\Theta_t) = T(B(H(x;\Theta_h);\Theta_b);\Theta_t)
\end{equation}

Without any change to head and tail, we can simply modify the number of repeated layers in the body to change the network capacity. We extract the intermediate feature ${f_i}$ of the body, where $i \in [1,N]$, to generate the early-exit output:
\begin{equation}
\widehat {{y_i}} = T({f_i};{\Theta_t})
\end{equation}

The original SR network uses the last layer's feature ${f_N}$ to generate the output. As the intermediate features of body have the same resolution as last layer's feature, we construct the multi-exit SR network by exit early in the intermediate layers. Different exits require different computational resources. We initialize the multi-exit SR network with the pre-trained model, and all exits' L1 losses are summed up as the multi-exit SR network's reconstruction loss $L_m$:
\begin{equation}
L_m = \sum_{i=1}^N |\hat{y_i} - y|
\label{eq:loss_multi}
\end{equation}

where ${y}$ represents the HR image and $N$ is the total number of layers. The training details are identical to the setup described in Sec. \ref{sec:train_setup}.

\subsection{Estimating Incremental Capacity}
\label{sec:estimate}

\begin{figure*}[t]
	\centering
	\begin{subfigure}{0.48\linewidth}
		\includegraphics[width=\textwidth]{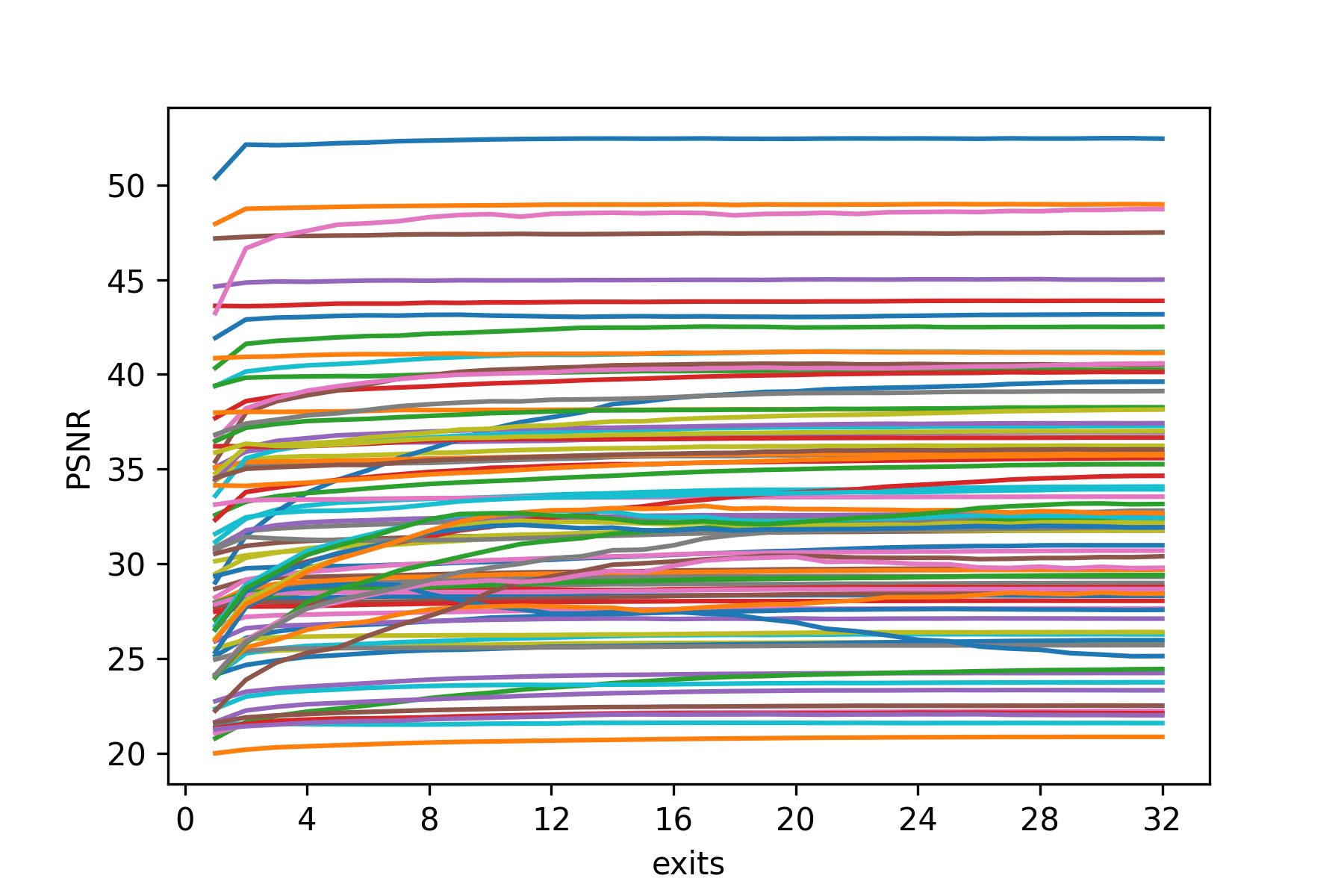}
		\caption{All Patches}
		\label{fig:patches_all}
	\end{subfigure}
	\begin{subfigure}{0.48\linewidth}
		\includegraphics[width=\textwidth]{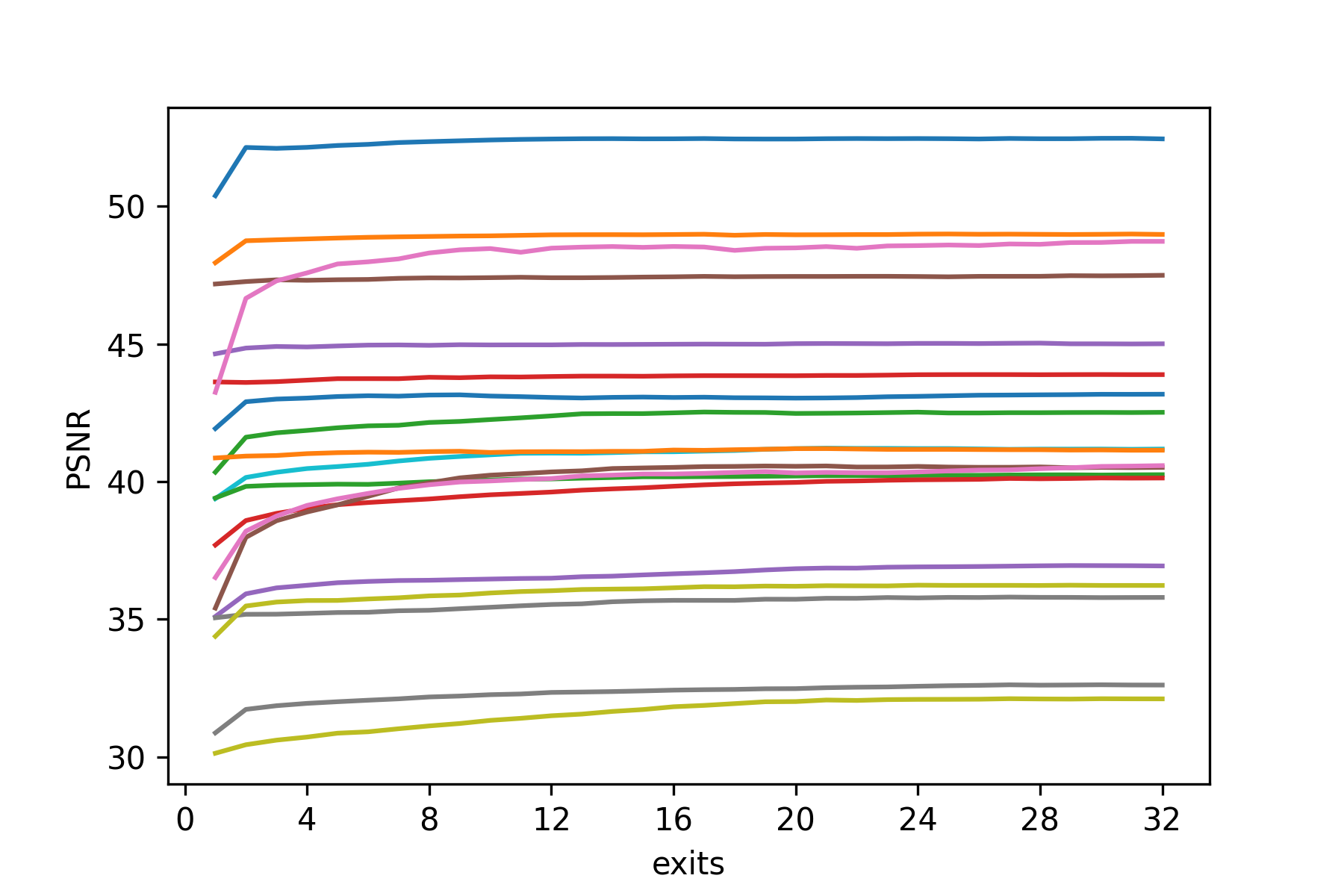}
		\caption{Bottleneck Patches}
		\label{fig:patches_bottleneck}
	\end{subfigure}
	\begin{subfigure}{0.48\linewidth}
		\includegraphics[width=\textwidth]{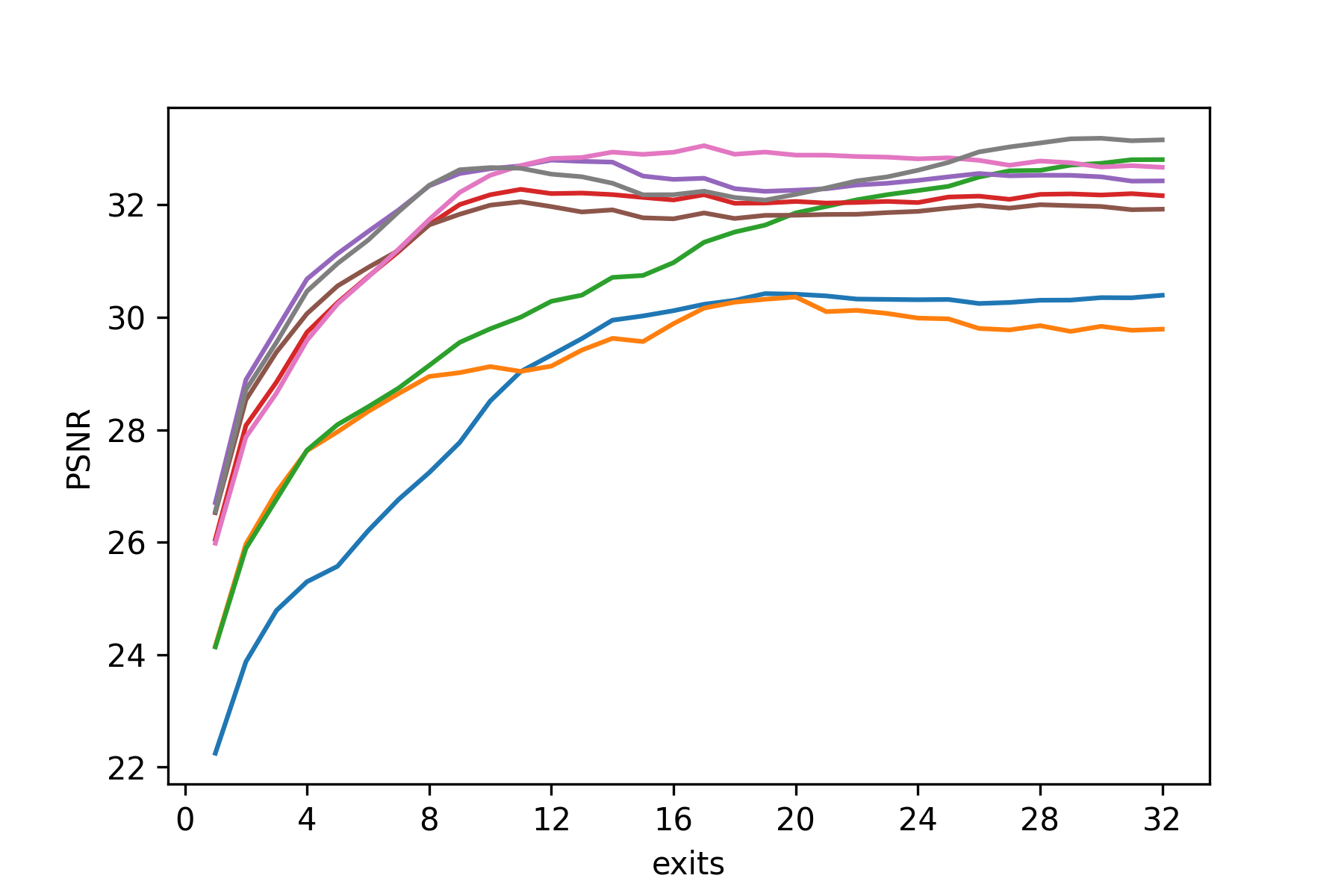}
		\caption{Growing Patches}
		\label{fig:patches_growing}
	\end{subfigure}
	\begin{subfigure}{0.48\linewidth}
		\includegraphics[width=\textwidth]{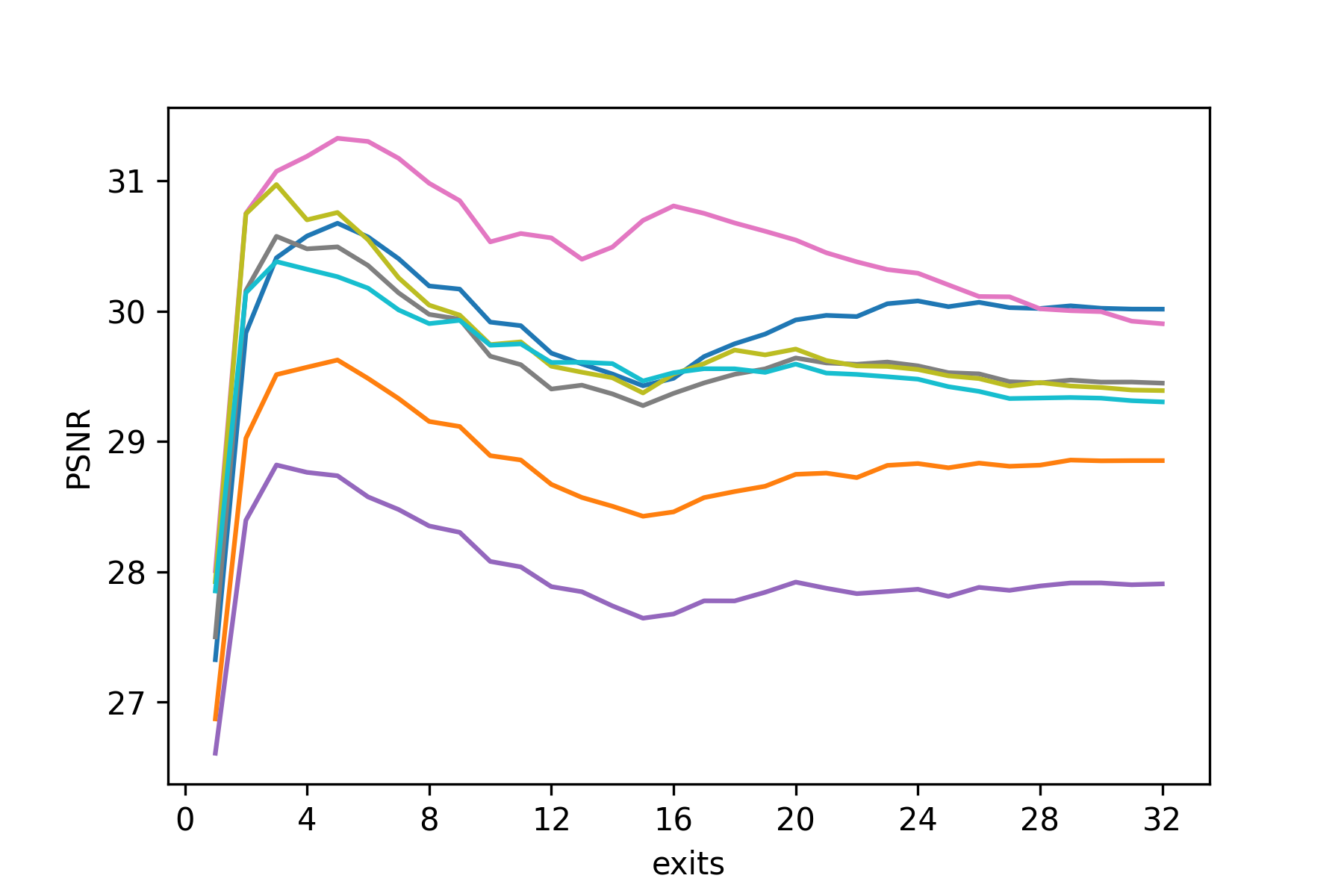}
		\caption{Overfitting Patches}
		\label{fig:patches_overthinking}
	\end{subfigure}
	\caption{Layer-wise PSNR performances of different patches. All patches are of size 32 × 32 and super-resolved by 32-exit EDSR at ×2 scaling factor. The layer-wise performances of all sampled patches are reported in (a). There are three types of patches according to our observation, which are called bottleneck patches, growing patches, overfitting patches respectively as shown in (b), (c) and (d)}
	\label{fig:samples}
\end{figure*}

In order to make the multi-exit SR networks scalable,
we need to design the signal of early-exit at a certain layer. Therefore, we train a 32-exit EDSR on DIV2K, and randomly sample 32 × 32 LR patches to observe their layer-wise performances. The result is shown in Fig. \ref{fig:patches_all}. A naive method of adaptive inference is to exit early when the performance is exceeding a threshold. However, we observe that there are three types of patches. The first one is named as bottleneck patches, which can achieve satisfying performance with a few layers as shown in Fig. \ref{fig:patches_bottleneck}. The second one is called growing patches, which need more layers to achieve good performance as shown in Fig. \ref{fig:patches_growing}. The third one is called over-fitting patches as shown in Fig. \ref{fig:patches_overthinking}, which might even achieve worse performance with more layers. In addition, the intervals of these three types are also quite different. The above observation shows that the signal of early-exit should be released when the performance gets saturated, rather than when the performance exceeding a threshold.

We define the early-exit signal $I_i$ as the incremental capacity of ${i^{th}}$ layer, which measures the performance difference between current layer and previous layer:
\begin{equation}
I_i = \sigma(P_i - P_{i-1})
\label{eq:incremental_capacity}
\end{equation}

where $\sigma$ is the tanh function, $P_i$ is the reconstruction performance of ${i^{th}}$ layer. As can be seen, the range of $I_i$ is $[-1, 1]$. Higher incremental capacity means more performance gain when forwarding ${i^{th}}$ layer. When $I_i$ is close to $0$, it means the performance get saturated. When $I_i$ is below $0$, it indicates the performance will get worse. In this paper, we use the PSNR between SR image and HR image as the reconstruction performance:
\begin{equation}
P_i = PSNR(\hat{y_i}, y)
\label{eq:psnr}
\end{equation}

\cite{kong2021classsr} proposes to train a module to classify the patches into different difficulties. However, as we have mentioned above, the relation between network capacity and performance is not monotonic. There are some patches that achieve worse results with more layers. Instead, using the incremental capacity can always correctly measure the saturation of performance and exit at the optimal layer.

During inference, since we cannot get the accurate incremental capacity due to the lack of HR image. Therefore, we propose to train a lightweight regressor $R$, which takes the feature $f_i$ of each layer in the body as input, to estimate the ${i^{th}}$ layer's incremental capacity $\hat{I_i}$. All the layers share the same regressor:
\begin{equation}
\hat{I_i} = \sigma(W \ast g(f_i) + b)
\label{eq:regressor}
\end{equation}

The regressor contains a fully-connected layer, where $g$ is global average pooling operation, $W$ and $b$ are the weight and bias of the fully-connected layer. The loss function of the regressor is the L2 loss between $\hat{I_i}$ and ground-truth incremental capacity $I_i$:
\begin{equation}
L_i = {\parallel}\hat{I_i} - I_i{\parallel}_2^2
\label{eq:l2}
\end{equation}

\subsection{Jointly Training SR Network and Regressor}
\label{sec:ape}

To apply APE to a SR network, we train its multi-exit SR network and the regressor jointly. The overall loss consists of all layers' reconstruction loss and the incremental capacity regression loss:
\begin{equation}
L = L_m + \lambda\sum_{i=1}^{N}L_i
\label{eq:all}
\end{equation}

where $\lambda$ is a hyper-parameter to balance these two losses, and we set it to 1 for all our experiments. During inference, we split the input image into overlapped patches, and feed all the patches into multi-exit SR network in parallel. Once the incremental capacity of a patch is below a given threshold, the patch can exit early. Increasing the threshold will make patches exit earlier and reduce the computational cost. Finally, the HR patches are merged to obtain the output image.

\section{Experiments}
\subsection{Implementation Details}

\subsubsection{Training Setup}
\label{sec:train_setup}
We use DIV2K dataset \cite{Agustsson_2017_CVPR_Workshops} to train all the models. The low-resolution images are generated by bicubic downsampling with scaling factors $\times2$, $\times3$ and $\times4$. Following former works, we use the first 800 images as the training set and 10 images (0801-0810) as the validation set. During training, data augmentation including random horizontal flip, random vertical flip and 90$^{\circ}$ rotation are applied. We train all the models for 300 epochs with learning rate initialed as 1e-4 and decayed to half at 200 epochs. The batch size is 16 and the HR patch size is 192. We use Adam optimizer, where $\beta_1$ is set to 0.9 and $\beta_2$ is set to 0.999.

\subsubsection{Testing Setup}
We use DIV2K \cite{Agustsson_2017_CVPR_Workshops} dataset and DIV8K \cite{gu2019div8k} dataset for testing since the widely-used benchmark datasets are not suitable for large image super-resolution evaluation. Specifically, we choose 100 images (0801-0900) from DIV2K for high-definition (HD) scenario, and 100 images (1401-1500) from DIV8K for ultra high-definition (UHD) scenario. During testing, we first split LR images into patches of size 48 with stride 46 unless otherwise specified. Then the LR patches are super-resolved in parallel, and the parallel size can be tuned to fit the computational resource. Finally, the SR patches are merged to obtain the complete SR images by weighting overlapping areas. The Peak Signal-to-Noise Ratio (PSNR) and Structural Similarity (SSIM) calculated on RGB channels are adopted as the evaluation metrics to measure super-resolution performance. We use FLOPs to evaluate the computational cost and the practical running time is benchmarked on NVIDIA 2080Ti GPUs.

\begin{table}[t]
	\centering
    \setlength{\tabcolsep}{3pt}
    \renewcommand{\arraystretch}{1.0}
	\caption{Performance evaluation of APE. FLOPs, PSNR and SSIM on DIV2K and DIV8K datasets with scaling factors $\times2$, $\times3$, $\times4$ are reported in the table. To compare the performance of APE with baselines, incremental capacity threshold is set to $0$. Therefore, all the patches can exit at the optimal layers.}
	\begin{tabular}{@{}lccccccc@{}}
		\toprule
		
		\multirow{2}*{Method} & \multirow{2}*{Scale} & \multicolumn{3}{c}{DIV2K}              &\multicolumn{3}{c}{DIV8K} \\
        \cmidrule(r){3-5}  \cmidrule(r){6-8}
		~ & ~ & FLOPS         & PSNR         & SSIM     & FLOPS           & PSNR         & SSIM  \\
		
		\toprule     
		     
		ECBSR     & $\times2$    & 1.38G         & 33.86dB      & 0.9309      & 1.38G           & 39.82dB      & 0.9649       \\
		ECBSR-APE & $\times2$    & 1.37G         & \textcolor{red}{33.87dB}      & \textcolor{red}{0.9316}      & 1.37G           & 39.73dB      & 0.9646       \\
		VDSR      & $\times2$    & 6.17G         & 33.63dB      & 0.9286      & 6.17G           & 39.71dB      & 0.9640       \\
		VDSR-APE  & $\times2$    & 6.14G         & 33.62dB      & \textcolor{red}{0.9292}      & 6.15G           & 39.54dB      & 0.9636       \\
		RCAN      & $\times2$    & 35.36G        & 34.09dB      & 0.9330      & 35.36G          & 40.04dB      & 0.9657       \\
		RCAN-APE  & $\times2$    & 35.36G        & \textcolor{red}{34.36dB}      & \textcolor{red}{0.9357}      & 35.36G          & \textcolor{red}{40.22dB}      & \textcolor{red}{0.9663}       \\
		EDSR      & $\times2$    & 93.89G        & 34.21dB      & 0.9343      & 93.89G          & 39.97dB      & 0.9656       \\
		EDSR-APE  & $\times2$    & 93.89G        & \textcolor{red}{34.46dB}      & \textcolor{red}{0.9366}      & 93.89G          & \textcolor{red}{40.16dB}      & \textcolor{red}{0.9662}       \\
     
		\midrule     
     
		ECBSR     & $\times3$    & 1.40G         & 30.22dB      & 0.8606      & 1.40G           & 35.36dB      & 0.9158       \\
		ECBSR-APE & $\times3$    & 1.40G         & 30.16dB      & \textcolor{red}{0.8618}      & 1.40G           & 35.31dB      & \textcolor{red}{0.9159}       \\
		VDSR      & $\times3$    & 13.88G        & 29.99dB      & 0.8567      & 13.88G          & 35.15dB      & 0.9133       \\
		VDSR-APE  & $\times3$    & 13.88G        & 29.92dB      & \textcolor{red}{0.8574}      & 13.88G          & \textcolor{red}{35.15dB}      & \textcolor{red}{0.9138}       \\
		RCAN      & $\times3$    & 35.80G        & 30.45dB      & 0.8645      & 35.80G          & 35.44dB      & 0.9171       \\
		RCAN-APE  & $\times3$    & 35.74G        & \textcolor{red}{30.59dB}      & \textcolor{red}{0.8695}      & 35.76G          & \textcolor{red}{35.65dB}      & \textcolor{red}{0.9192}       \\
		EDSR      & $\times3$    & 100.77G       & 30.55dB      & 0.8669      & 100.77G         & 35.54dB      & 0.9178       \\
		EDSR-APE  & $\times3$    & 100.77G       & \textcolor{red}{30.66dB}      & \textcolor{red}{0.8711}      & 100.77G         & \textcolor{red}{35.68dB}      & \textcolor{red}{0.9197}       \\
     
		\midrule     
     
		ECBSR     & $\times4$    & 1.43G         & 28.29dB      & 0.8026      & 1.43G           & 33.07dB      & 0.8724       \\
		ECBSR-APE & $\times4$    & 1.43G         & \textcolor{red}{28.31dB}      & \textcolor{red}{0.8036}      & 1.43G           & 33.05dB      & 0.8719       \\
		VDSR      & $\times4$    & 24.68G        & 28.12dB      & 0.7974      & 24.68G          & 32.88dB      & 0.8688       \\
		VDSR-APE  & $\times4$    & 24.53G        & 28.10dB      & \textcolor{red}{0.7991}      & 24.53G          & 32.83dB      & \textcolor{red}{0.8689}       \\
		RCAN      & $\times4$    & 36.77G        & 28.52dB      & 0.8077      & 36.77G          & 33.25dB      & 0.8753       \\
		RCAN-APE  & $\times4$    & 36.71G        & \textcolor{red}{28.70dB}      & \textcolor{red}{0.8138}      & 36.74G          & \textcolor{red}{33.38dB}      & \textcolor{red}{0.8776}       \\
		EDSR      & $\times4$    & 115.83G       & 28.66dB      & 0.8112      & 115.83G         & 33.30dB      & 0.8762       \\
		EDSR-APE  & $\times4$    & 115.79G       & \textcolor{red}{28.78dB}      & \textcolor{red}{0.8158}      & 115.81G         & \textcolor{red}{33.39dB}      & \textcolor{red}{0.8779}       \\
		
		\bottomrule
	\end{tabular}
	\label{tab:performance}
\end{table}

\begin{table}[t]
	\centering
    \setlength{\tabcolsep}{3pt}
    \renewcommand{\arraystretch}{1.0}
	\caption{Efficiency evaluation of APE under the same performance as original SR networks. Parameters, body FLOPs, total FLOPs, and practical inference times on DIV2K with scaling factors $\times2$, $\times3$, $\times4$ are reported in the table. Inference time is evaluated on NVIDIA 2080Ti GPUs.}
	\begin{tabular}{@{}lcccclcc@{}}
		\toprule

		Method        & Scale        & Param.   & PSNR         & Body FLOPs         & Total FLOPs      & Time (ms)   \\
		\toprule    
		    
		ECBSR         & $\times2$    & 1.0K         & 33.86dB      &  1.36G             & 1.38G (100\%)    & 244          \\
		ECBSR-APE     & $\times2$    & 1.0K         & 33.82dB      &  0.99G   & 1.01G (73\%)     & 211          \\
		VDSR          & $\times2$    & 0.67M        & 33.63dB      &  6.14G             & 6.17G (100\%)    & 346          \\
		VDSR-APE      & $\times2$    & 0.67M        & 33.61dB      &  4.13G     & 4.16G (67\%)     & 334          \\
		RCAN          & $\times2$    & 15.4M        & 34.09dB      &  34.91G            & 35.36G (100\%)   & 2323         \\
		RCAN-APE      & $\times2$    & 15.4M        & 34.09dB      &  8.26G    & 8.71G (24\%)     & 974          \\
		EDSR          & $\times2$    & 40.7M        & 34.21dB      &  87.01G            & 93.89G (100\%)   & 2133         \\
		EDSR-APE      & $\times2$    & 40.7M        & 34.21dB      &  34.07G    & 40.95G (43\%)    & 733          \\

		\midrule

		ECBSR         & $\times3$    & 1.0K         & 30.21dB      &  1.36G             & 1.40G (100\%)    & 239          \\
		ECBSR-APE     & $\times3$    & 1.0K         & 30.12dB      &  1.01G   & 1.05G (74\%)     & 219          \\
		VDSR          & $\times3$    & 0.67M        & 29.99dB      &  13.80G            & 13.88G (100\%)   & 346          \\
		VDSR-APE      & $\times3$    & 0.67M        & 29.91dB      &  11.01G    & 11.09G (79\%)    & 325          \\
		RCAN          & $\times3$    & 15.6M        & 30.45dB      &  34.91G            & 35.80G (100\%)   & 1040         \\
		RCAN-APE      & $\times3$    & 15.6M        & 30.45dB      &  13.57G   & 14.46G (40\%)    & 627          \\
		EDSR          & $\times3$    & 43.7M        & 30.55dB      &  87.01G            & 100.77G (100\%)  & 1777         \\
		EDSR-APE      & $\times3$    & 43.7M        & 30.55dB      &  49.28G   & 63.04G (62\%)    & 492          \\
    
		\midrule    
    
		ECBSR         & $\times4$    & 1.0K         & 28.29dB      &  1.36G             & 1.43G (100\%)    & 245          \\
		ECBSR-APE     & $\times4$    & 1.0K         & 28.22dB      &  0.85G   & 0.92G (64\%)     & 196          \\
		VDSR          & $\times4$    & 0.67M        & 28.12dB      &  24.55G            & 24.68G (100\%)   & 345          \\
		VDSR-APE      & $\times4$    & 0.67M        & 28.07dB      &  17.75G    & 17.88G (72\%)    & 303          \\
		RCAN          & $\times4$    & 15.6M        & 28.53dB      &  34.91G            & 36.77G (100\%)   & 620          \\
		RCAN-APE      & $\times4$    & 15.6M        & 28.53dB      &  10.53G   & 12.39G (33\%)    & 350          \\
		EDSR          & $\times4$    & 43.1M        & 28.66dB      &  87.01G            & 115.83G (100\%)  & 1123         \\
		EDSR-APE      & $\times4$    & 43.1M        & 28.67dB      &  49.86G    & 78.68G (67\%)    & 419          \\
		
		\bottomrule
	\end{tabular}
	\label{tab:efficiency}
\end{table}

\subsection{Evaluation of APE}
\subsubsection{Performance Results}
To evaluate the effectiveness of our method, we apply APE to state-of-the-art SR networks, including ECBSR \cite{zhang2021edge}, VDSR \cite{kim2016accurate}, EDSR \cite{lim2017enhanced} and RCAN \cite{zhang2018image}. We set the threshold of incremental capacity to 0 and evaluate the performance of APE on DIV2K and DIV8K. As we have mentioned above, there are three types of patches. The over-fitting patches might achieve worse performance with more layers. Therefore, setting the threshold to 0 enables all the patches to exit at the optimal layers. As shown in Tab. \ref{tab:performance}, SR networks with APE can achieve comparable or even superior performance compared to original SR networks in terms of PSNR and SSIM. This comparison demonstrates that incremental capacity is a more reasonable metric to evaluate the contribution of each layer.

\subsubsection{Efficiency Results}
As for the efficiency of APE, Tab. \ref{tab:efficiency} shows the detailed computational cost under same performance as original SR networks. Our method adds a lightweight regressor whose FLOPs is negligible. APE can significantly reduce the computational cost of original SR networks across different scaling factors. For example, RCAN-APE only needs 24\%, 40\%, and 33\% of original computational cost on scaling factors $\times2$, $\times3$ and $\times4$. The computational cost of body is significantly reduced by our method, and the computational costs of head and tail stay the same. Overall, our method can nearly halve the computational cost under the same performance. 

\subsubsection{Scalability Results}
\begin{figure*}[t]
	\centering
	\begin{subfigure}{0.49\linewidth}
		\includegraphics[width=\textwidth]{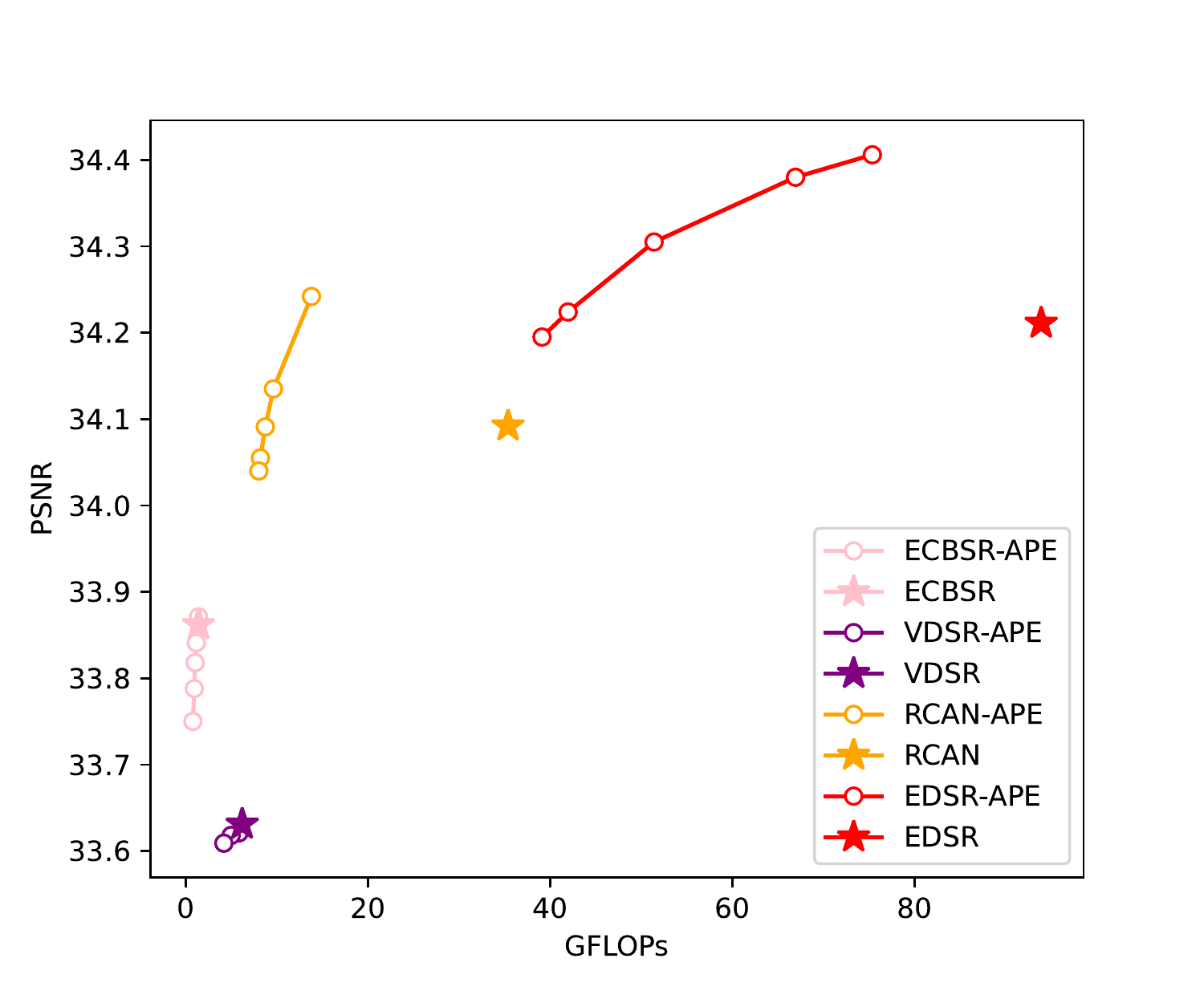}
		\caption{PSNR-GFLOPs on DIV2K $\times2$}
		\label{fig:div2k_x2_psnr}
	\end{subfigure}
	\begin{subfigure}{0.49\linewidth}
		\includegraphics[width=\textwidth]{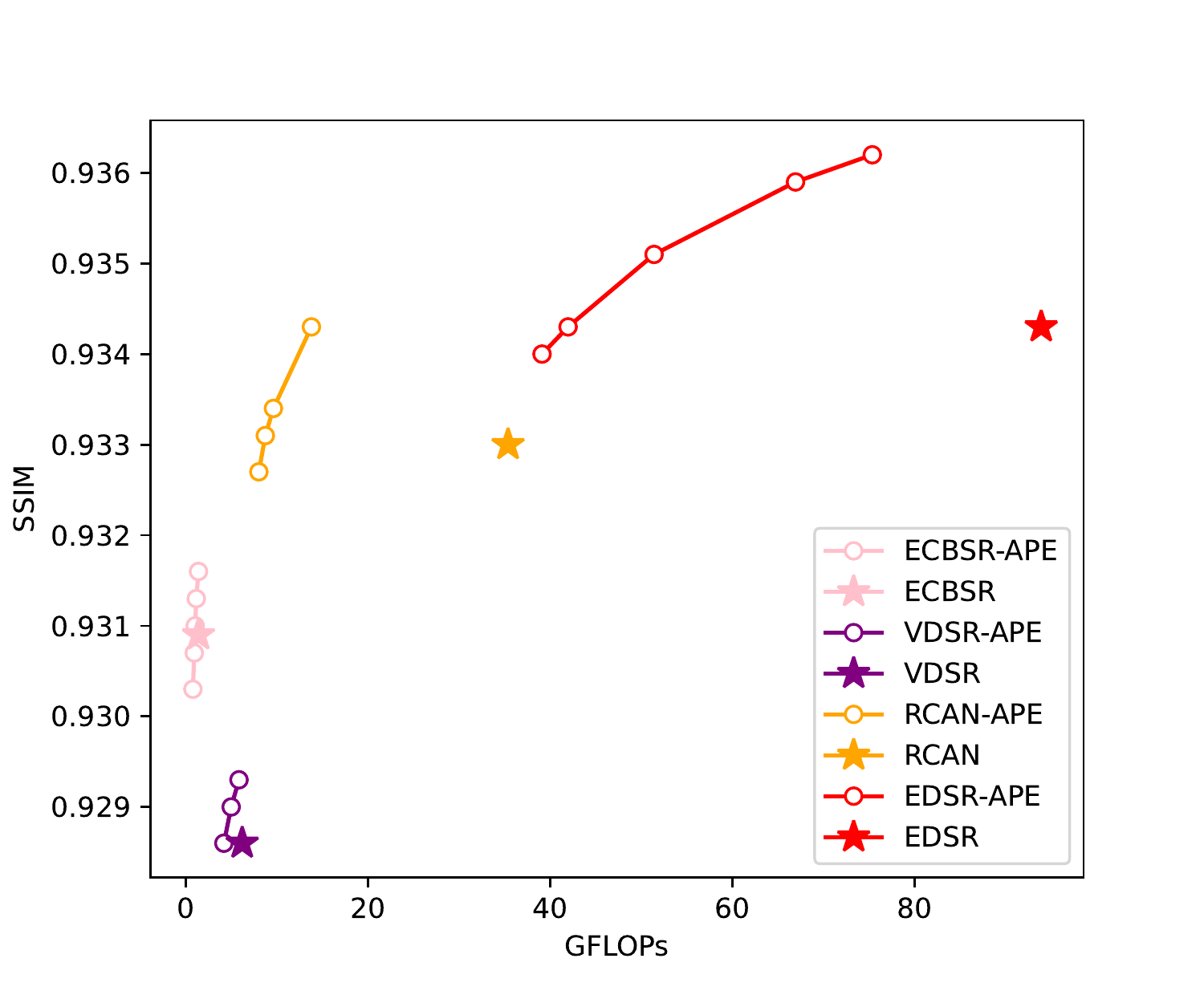}
		\caption{SSIM-GFLOPs on DIV2K $\times2$}
		\label{fig:div2k_x2_ssim}
	\end{subfigure}
	\begin{subfigure}{0.49\linewidth}
		\includegraphics[width=\textwidth]{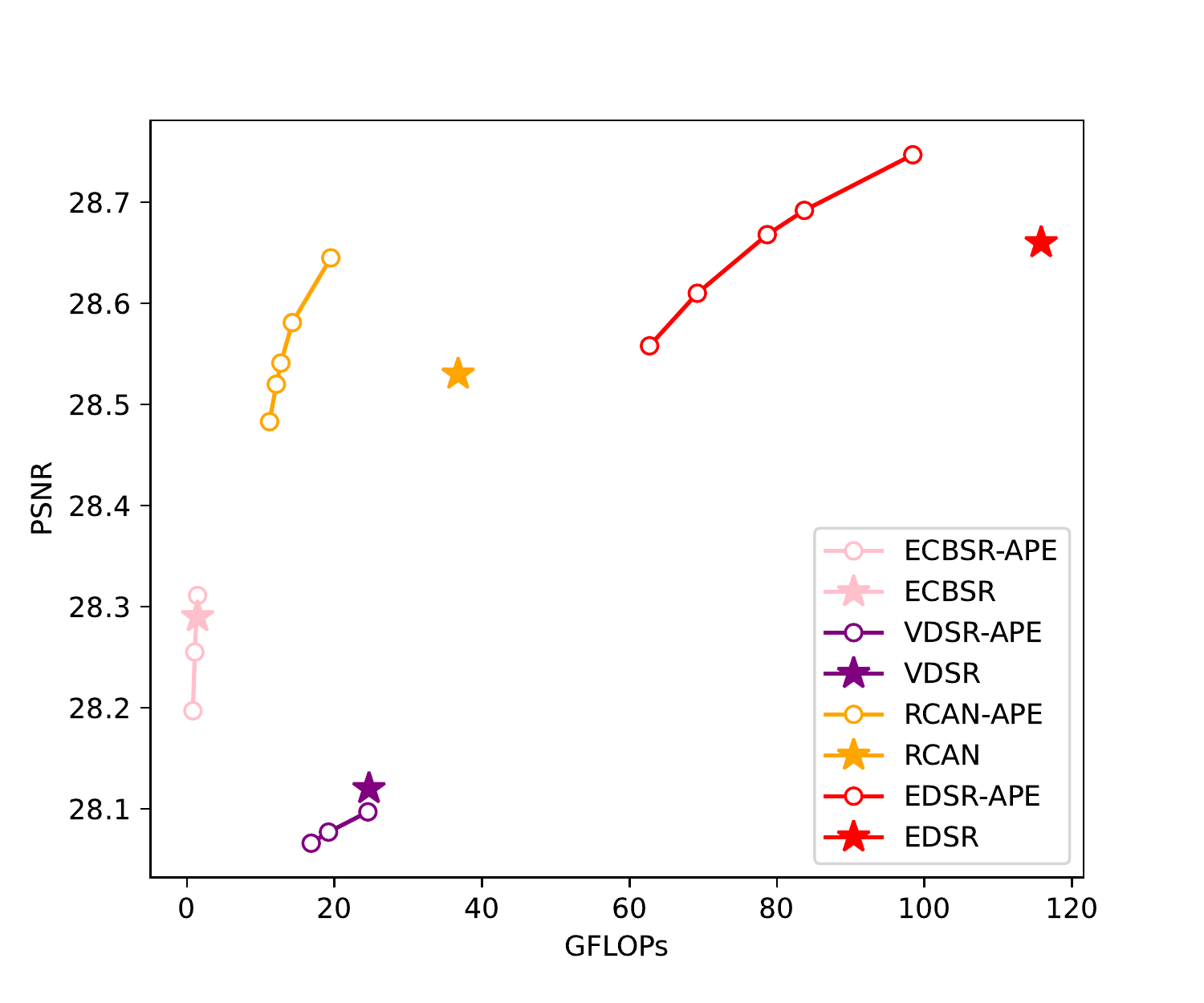}
		\caption{PSNR-GFLOPs on DIV2K $\times4$}
		\label{fig:div2k_x4_psnr}
	\end{subfigure}
	\begin{subfigure}{0.49\linewidth}
		\includegraphics[width=\textwidth]{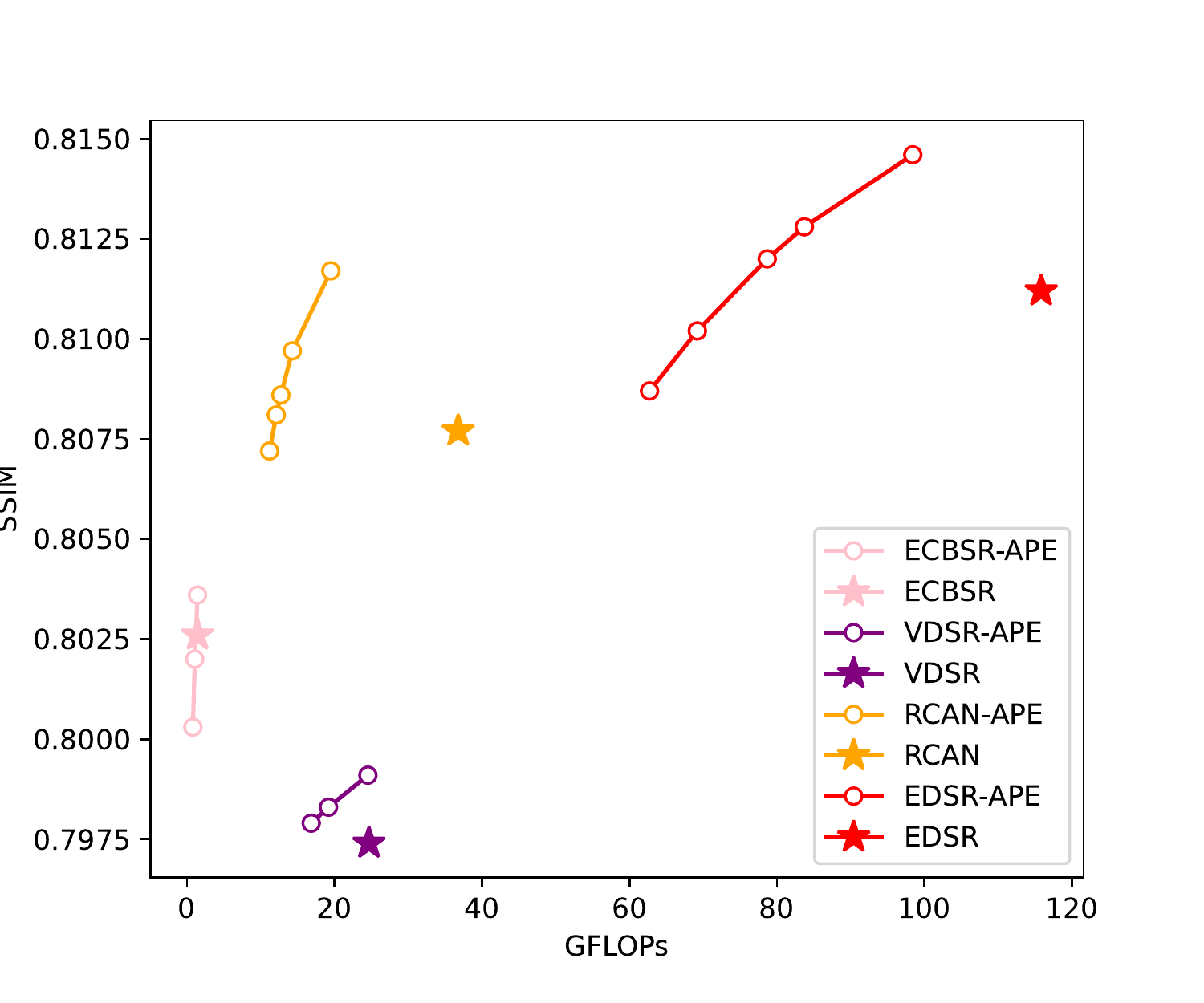}
		\caption{SSIM-GFLOPs on DIV2K $\times4$}
		\label{fig:div2k_x4_ssim}
	\end{subfigure}
	\caption{Quantitative results of performance-efficiency trade-off. We apply APE to ECBSR, VDSR, EDSR and RCAN with scaling factors $\times2$ and $\times4$ on DIV2K dataset. Average FLOPs of all 48$\times$48 LR patches and PSNR/SSIM calculated on the full image are reported.}
	\label{fig:trade-off}
\end{figure*}

We also show the performance-efficiency trade-off results in Fig. \ref{fig:trade-off} to demonstrate the scalability of APE. By controlling the incremental capacity threshold, APE can achieve scalable performance-efficiency trade-off. Therefore, we can deploy one APE SR network on platforms with different computational resources. For the device with low computational resource, we can raise the threshold to get lower performance and faster inference speed.

\subsubsection{Visual Results}
\begin{figure}[t]
	\centering
	\includegraphics[width=\textwidth]{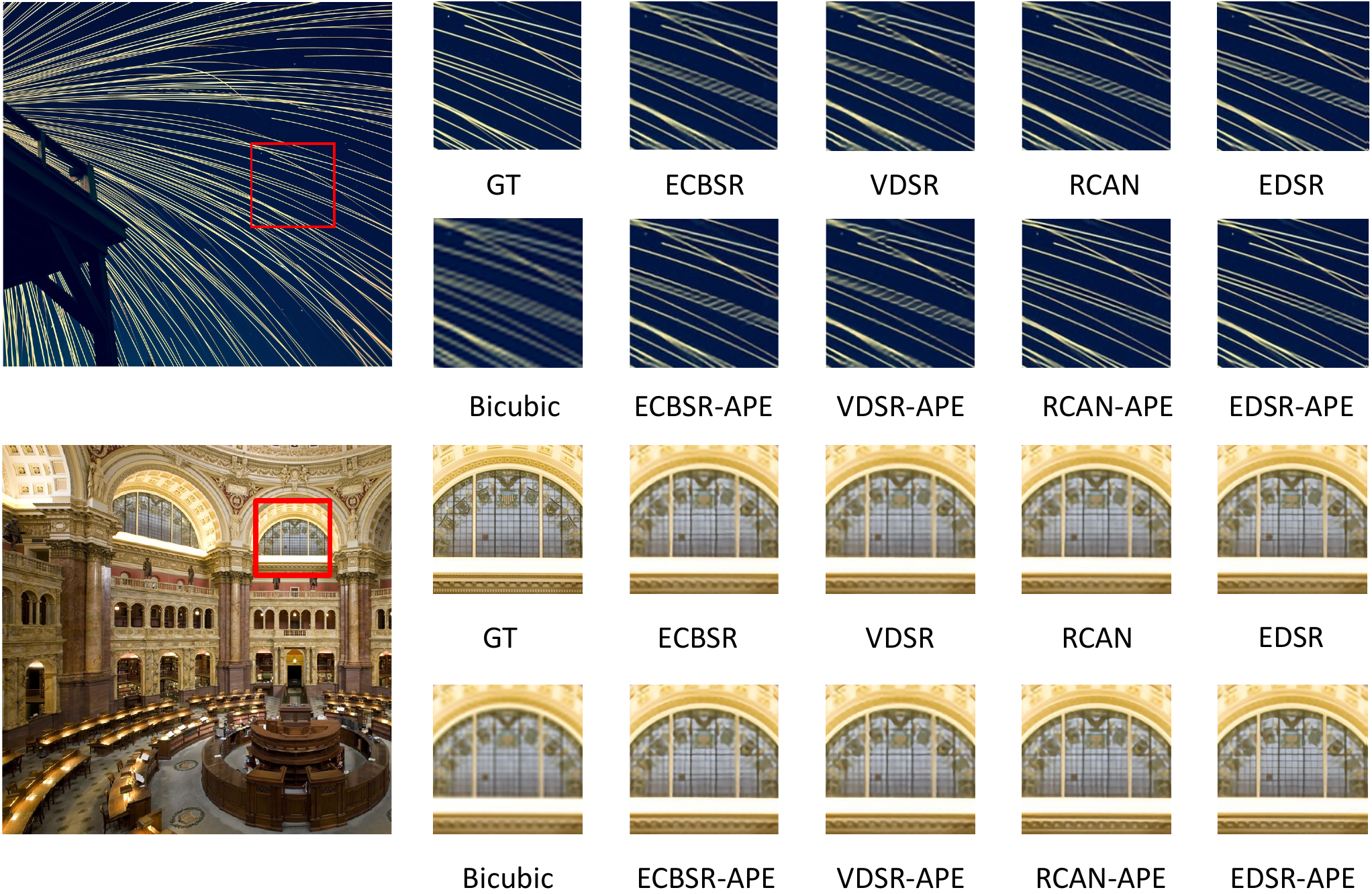}
	\caption{Qualitative results of APE and original SR networks on DIV2K dataset with $\times4$ scaling factor.}
	\label{fig:visual}
\end{figure}

Fig. \ref{fig:visual} shows the qualitative comparison of our method against the original SR networks. As we can see, EDSR-APE and RCAN-APE can achieve same or even better visual results compared with original SR networks. Although we merge the patches to obtain complete SR images, weighting overlapped patches can avoid the stitching artifacts.

\begin{figure}
	\centering
	\begin{subfigure}{0.49\linewidth}
		\includegraphics[width=\textwidth]{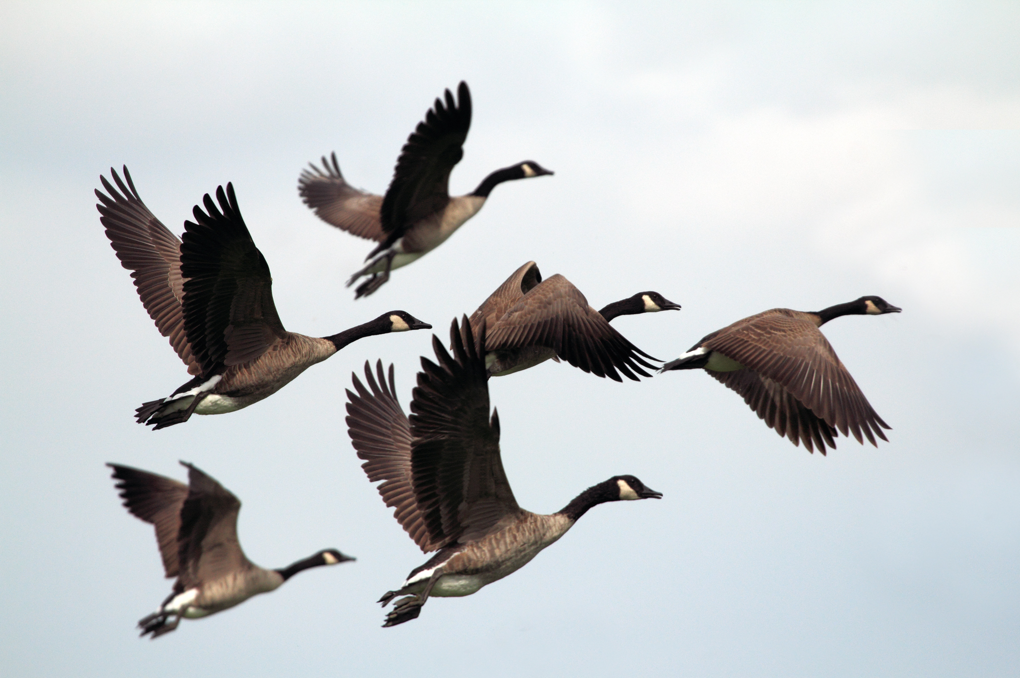}
		\caption{Original image}
	\end{subfigure}
	\hfill
	\begin{subfigure}{0.49\linewidth}
		\includegraphics[width=\textwidth]{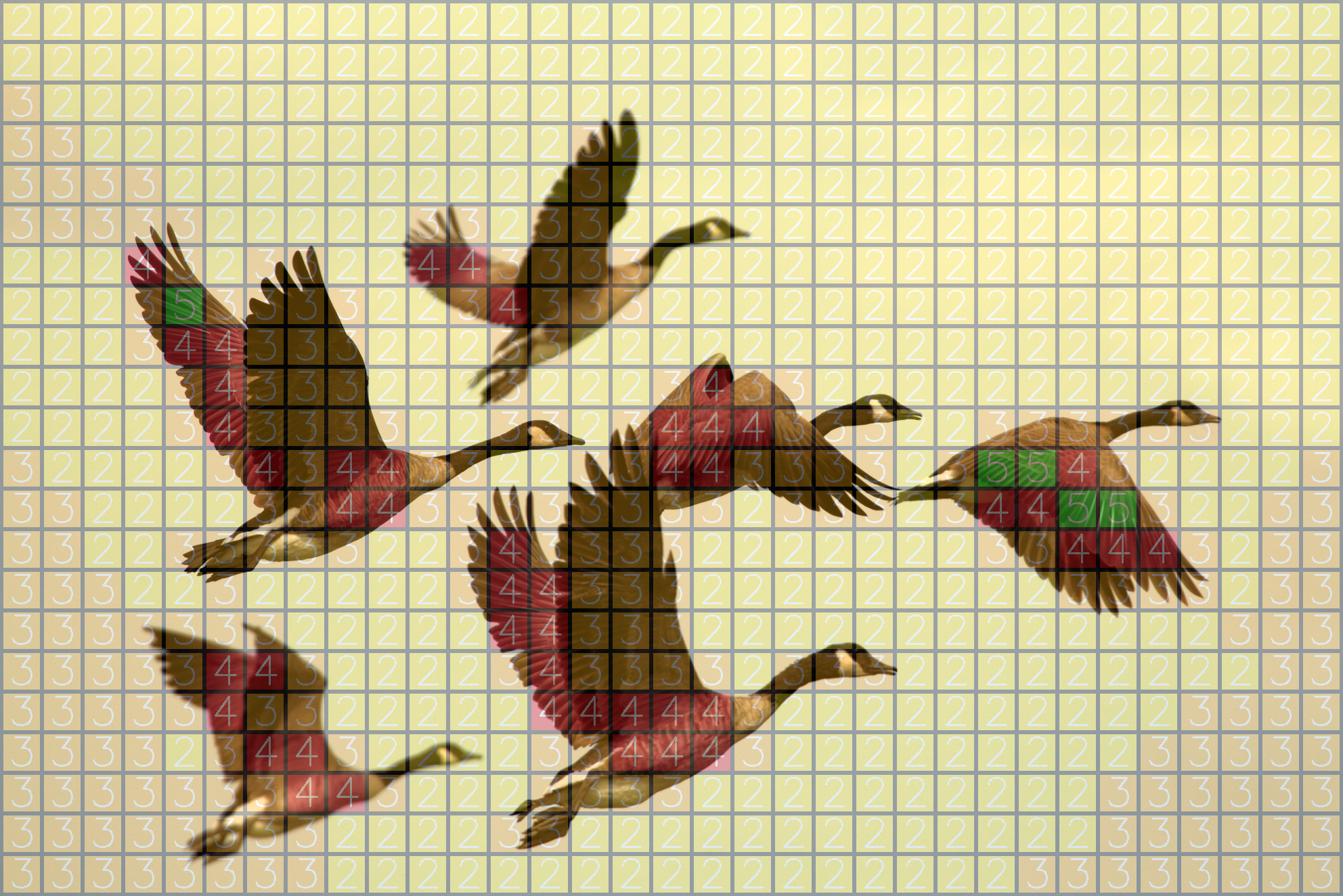}
		\caption{Adaptive exiting patches}
	\end{subfigure}
	\caption{Visualization of early-exit patches. The number in the patch indicates the exit index of each patch. Best viewed by zooming x4.}
	\label{fig:patches}
\end{figure}

Fig. \ref{fig:patches} visualizes the status of adaptive exiting patches. As can be seen, most patches in smooth regions exit at the early layers since they are easy to be restored. As for patches in complicated regions, they will exit at the later layers. This is consistent with the motivation of applying appropriate networks to various difficulties.

\subsection{Ablation Study}
\subsubsection{Variants of Exit Interval}
\label{sec:exit_interval}
\begin{figure}[t]
    \centering
	\begin{minipage}{0.48\linewidth}
	\centering
		\includegraphics[width=\textwidth]{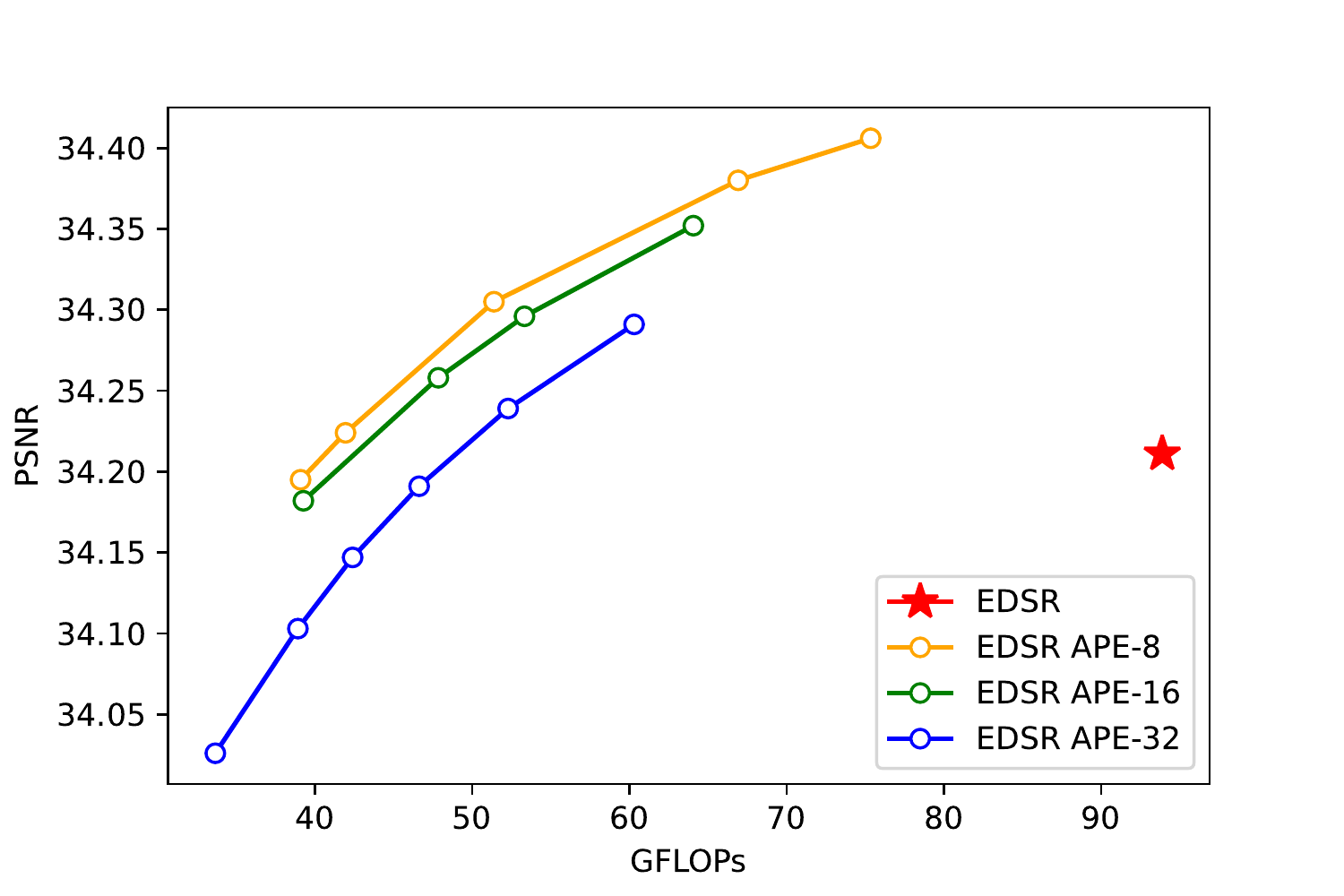}
		\caption{Variants of exit interval. We show the results of different exit intervals by evaluating EDSR-APE on the DIV2K dataset with scaling factors $\times2$. APE-$n$ means APE with $n$ exits.}
		\label{fig:interval}
	\end{minipage}%
	\hfill
	\begin{minipage}{0.48\linewidth}
	\centering
		\includegraphics[width=\textwidth]{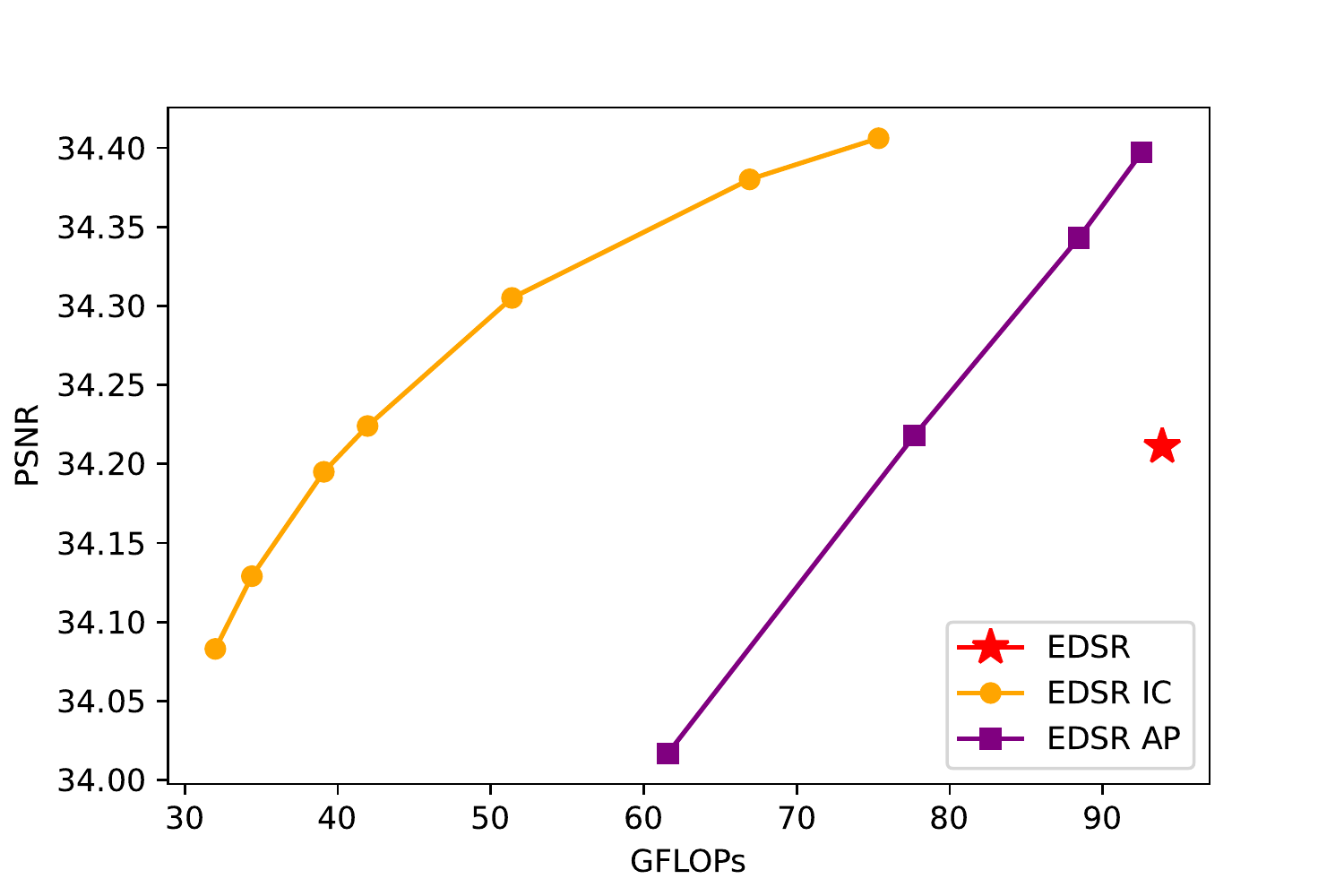}
		\caption{Variants of early-exit signal. $IC$ denotes incremental capacity, and $AP$ denotes absolute performance. Both are evaluated on DIV2K dataset with $\times2$ scaling factor.}
	    \label{fig:signal}
	\end{minipage}
\end{figure}

We conduct the experiment to study the influence of different exit intervals using EDSR as the backbone. Specifically, EDSR has 32 repeated residual layers in the body. We evaluate the exit intervals of 4, 2 and 1 layers. The results on DIV2K dataset with scaling factors $\times2$ are shown in Fig. \ref{fig:interval}. As can be seen, exit interval of 4 layers achieves the best results, indicating that multi-exit SR networks need sufficient learning capacity within each exit.

\subsubsection{Variants of Early-Exit Signal}
\label{sec:signal}
Apart from the proposed incremental capacity ($IC$), we can also use the absolute performance ($AP$) of each layer as the early-exit signal to measure the necessity of early-exiting at specific layer. We compare the results of incremental capacity and absolute performance in Fig. \ref{fig:signal}. As can be seen, compared with absolute performance, incremental capacity can reduce more computational cost, validating that incremental capacity is the better early-exit signal for multi-exit SR networks.

\subsubsection{Variants of Patch Size and Stride}
\label{sec:patch}
Since our method splits an image into patches, we evaluate the performance of different patch sizes and strides. As shown in Tab. \ref{tab:patch}, different patch size can achieve similar performance in terms of PSNR and SSIM.

\subsection{Comparison with ClassSR and AdaDSR}
We also compare with ClassSR \cite{kong2021classsr} and AdaDSR \cite{liu2020deep} on DIV2K $\times4$ using RCAN as the backbone under the same performance in Tab. \ref{tab:other}. We use the published codes of AdaDSR and ClassSR to perform the comparison. ClassSR \cite{kong2021classsr} is a patch-based SR method. It manually designs easy, medium and hard networks by changing the number of channels. A module is trained to classify the patches into easy, medium and hard. ClassSR \cite{kong2021classsr} can reduce the overall computational cost by applying different networks to different patches. However, they need to store all the models with different capacities, heavily increasing the model size. Apart from this, ClassSR classifies the patches into certain restoration difficulties. Therefore, it is not scalable over different computational resources. Instead, our method can easily adjust the trade-off between performance and efficiency to meet different computational resources. 

\begin{table}[t]
    \begin{minipage}[t]{0.48\linewidth}
	\centering
	\setlength{\tabcolsep}{3pt}
	\renewcommand{\arraystretch}{1.0}
	\caption{Variants of patch size and stride. PSNR and SSIM are evaluated on DIV2K dataset with scaling factor $\times4$. The numbers in the Patch column indicate (patch size, patch stride).}
	\begin{tabular}{@{}lccccc@{}}
		\toprule
		Method   & Patch      & PSNR         & SSIM     \\
		\toprule
		EDSR     & -          & 28.66        & 0.8112   \\
		EDSR-APE & (32,30)    & 28.702       & 0.8136   \\
		EDSR-APE & (40,38)    & 28.723       & 0.8151   \\
		EDSR-APE & (48,46)    & 28.783       & 0.8158   \\
		\bottomrule
	\end{tabular}
	\label{tab:patch}
	\end{minipage}%
	\hfill
	\begin{minipage}[t]{0.48\linewidth}
	\centering
	\setlength{\tabcolsep}{3pt}
	\renewcommand{\arraystretch}{1.0}
	\caption{Comparison with AdaDSR and ClassSR, APE achieves fastest inference speed without increasing the model's size. Besides, APE is scalable to different computational resources.}
	\begin{tabular}{@{}lccc@{}}
		\toprule
		Method       & Param.     & PSNR     & Time      \\
		\toprule
		RCAN         & 15.6M      & 28.526   & 620ms     \\
		RCAN-APE     & 15.6M      & 28.530   & 350ms     \\
		RCAN-Ada     & 15.7M      & 28.535   & 1644ms    \\
		RCAN-ClassSR & 30.1M      & 28.533   & 22s       \\
		\bottomrule
	\end{tabular}
	\label{tab:other}
	\end{minipage}
\end{table}

We also compare with AdaDSR \cite{liu2020deep}, which is based on pixel-wise sparse convolution. It will generate a spatially sparse mask for each layer and sparse convolution is conducted to achieve speedup. However, pixel-wise sparse convolution is not hardware-friendly on modern GPUs, thus there exists a gap between theoretical and practical speedup. As can be seen in Tab. \ref{tab:other}, with similar model size, APE is faster than AdaDSR in practice.

\section{Future Work}
Although our method can decide the optimal exit for each patch, we still rely on overlapped patches to avoid the stitching artifacts. Therefore, we can further improve the efficiency by adopting non-overlapped patches. Besides, we uniformly split an image into patches, which might not be the optimal solution for image splitting. Finally, applying our method to other low-level vision tasks is also a promising future work.

\section{Conclusion}
In this paper, we present adaptive patch exiting (APE) for scalable single image super-resolution. Since image patches are structured and have different restoration difficulties, we split an image into patches and train a regressor to predict the incremental capacity of each layer for the input patch. Therefore, the patch can exit at any layer by adjusting the threshold. We also propose a novel joint training strategy to train both the SR network and regressor. Extensive comparisons are conducted across various SR backbones, datasets and scaling factors to demonstrate the effectiveness of our method.

\clearpage
%
%
\bibliographystyle{splncs04}
\bibliography{egbib}
\end{document}